\documentclass[12pt,a4paper]{article}

\usepackage[utf8]{inputenc}
\usepackage[T1]{fontenc}
\usepackage{lmodern}
\usepackage[margin=2.5cm]{geometry}
\usepackage{graphicx}
\usepackage[export]{adjustbox}
\usepackage{amsmath,amssymb}
\usepackage{hyperref}
\usepackage{xcolor}
\usepackage{booktabs}
\usepackage{float}
\usepackage{caption}
\usepackage{subcaption}
\usepackage{setspace}
\usepackage{titlesec}
\usepackage{lineno}
\usepackage{microtype}
\usepackage[authoryear,round,sort]{natbib}
\hypersetup{
    colorlinks=true,
    linkcolor=blue!70!black,
    citecolor=blue!70!black,
    urlcolor=blue!70!black
}

\titleformat{\section}{\large\bfseries}{\thesection}{1em}{}
\titleformat{\subsection}{\normalsize\bfseries}{\thesubsection}{1em}{}

\title{\textbf{Central Dogma Transformer II}\\[0.3cm]
\large An AI Microscope for Understanding Cellular Regulatory Mechanisms}

\author{Nobuyuki Ota\footnote{Correspondence: \texttt{nobuyuki.ohta@gmail.com}}\\[0.2cm]
\small Independent Researcher, Burlingame, CA, USA}

\date{}

\begin{document}

\maketitle

\begin{abstract}
\noindent
\textbf{Motivation:} Interpretability is not optional in biology: understanding gene regulation requires models whose learned structure can be directly interrogated, not merely accurate predictors whose internals resist mapping onto regulatory relationships. We ask whether an architecture mirroring the central dogma yields attention and gradient maps that recover known regulatory elements and networks in inspectable form.

\textbf{Results:} Central Dogma Transformer II (CDT-II) mirrors the central dogma in its architecture---DNA self-attention, RNA self-attention, and DNA-to-RNA cross-attention---requiring only genomic embeddings and raw per-cell expression. On K562 CRISPR interference (CRISPRi) data with five genes held out entirely, CDT-II predicts perturbation effects (per-gene mean $r = 0.84$), recovers the \textit{GFI1B} regulatory network (6.6-fold enrichment, $P = 3.5 \times 10^{-17}$), and concentrates cross-attention on ENCODE regulatory elements including CTCF sites (mean 7.67$\times$ across 28 target genes, $P < 0.001$). Gradient attribution predicts consequences of perturbing therapeutic targets (mean $r = 0.82$). For \textit{TFRC}, target of the anti-TfR1 antibody PPMX-T003, it identifies erythrocyte-structure, iron-dependent DNA-synthesis and oxidative-stress genes, matching anemia and ferroptosis reported clinically and preclinically---without clinical data as input. CDT-II acts as an AI microscope, surfacing clinically relevant regulatory structure from perturbation experiments alone.

\textbf{Availability:} Source code is available at \url{https://github.com/nobusama/CDT2}. Pre-computed embeddings, training data, and model weights are available at \url{https://huggingface.co/datasets/nobusama17/CDT2-data}.

\textbf{Contact:} nobuyuki.ohta@gmail.com
\end{abstract}

\vspace{0.3cm}

\section{Introduction}

The central dogma \citep{crick1970} describes cellular information flow across three molecular layers: DNA encodes genetic instructions, RNA transmits and regulates this information, and proteins execute cellular functions. Artificial intelligence has transformed our ability to model each layer individually \citep{jumper2021,avsec2021enformer,cui2024,theodoris2023}, yet these models inherit what has become a defining challenge for large models across domains: a lack of interpretability \citep{novakovsky2023,eraslan2019}. Their internal representations often do not correspond directly to biological entities or relationships that researchers can examine and validate. In many applications, opacity is an acceptable price for accuracy; in biology it is not, because mechanistic understanding---knowing why a cell responds as it does---is the ultimate goal rather than a by-product, and it carries direct clinical implications, from anticipating drug side effects to identifying new therapeutic targets.

This is the black-box problem of biological AI. A model can predict a cellular outcome---whether a cell is cancerous, or how it will respond to a drug---yet reveal little about the regulatory events inside the cell that produce it. We argue that biology needs not only \textit{task-oriented} AI that optimizes such predictions, but also \textit{mechanism-oriented} AI: models that expose interpretable regulatory structure a researcher can inspect and test at the bench. Concretely, the questions that matter are which genomic regulatory elements control which genes, and how a perturbation at one locus propagates through the transcriptome---questions answered not by a single number but by interpretable structure. What we ultimately want is therefore an AI that, in the course of learning to predict, exposes the cell's regulatory machinery. We cast perturbation response as a prediction task: given a perturbed genomic locus and a cell's raw expression, predict the expression changes of genes held out from training. This prediction is the means, not the end. A model whose architecture mirrors the central dogma solves it largely by internalizing the regulatory structure that links locus to transcriptome, and that structure---not the prediction---is what we read out and test at the bench. Predictive accuracy is therefore evidence that the learned mechanism is faithful, not an end in itself. Such a model advances biology not by ranking higher on a shared prediction task, but by helping make the cell's regulatory interior visible where black-box predictors leave it hidden. We test this hypothesis by training on single-cell CRISPRi perturbation data and interrogating the trained model through its attention maps and gradient-based attribution.

To implement this principle, the Central Dogma Transformer \citep{ota2026cdtv1} mirrors the central dogma in its architecture---an ``AI microscope'' whose attention maps are directly interpretable as regulatory structure: DNA self-attention captures genomic relationships, RNA self-attention reflects gene co-regulation, and DNA-to-RNA cross-attention models transcriptional control. Crucially, the architecture reproduces not only the three molecular layers but the \textit{directionality} of information flow between them: the cross-attention is asymmetric by construction---RNA queries reading the genomic context---modeling a directed DNA-to-RNA regulatory relationship rather than a symmetric correlation between modalities. Because each attention mechanism corresponds to a specific biological relationship, the resulting maps provide direct readouts of regulatory organization. Interpretability here refers to CDT-II's own learned structure: its attention and gradient maps expose which regulatory elements and genes the model relies on, independent of whether the upstream sequence embeddings it consumes are themselves interpretable.

Equally important, CDT-II's success stems not only from model architecture but also from task formulation. The model receives raw per-cell expression values as input and predicts expression \textit{changes} (log2 fold changes) as output---without being provided the difference or any reference to the unperturbed state. This forces the model to learn what constitutes ``change'' and which genes influence which: to predict how gene B changes when locus A is perturbed, the model must internalize the regulatory relationship between A and B.

CDT-II builds on the foundation laid by CDT-I \citep{ota2026cdtv1}, which we summarize here to keep the present paper self-contained. CDT-I introduced the central-dogma architecture as a proof of concept spanning all three molecular layers: it combined three pre-trained language models---for DNA, RNA, and protein---into fixed, gene-level (non-cell-specific) embeddings linked by directional cross-attention, and validated the design on K562 CRISPRi enhancer perturbations, reaching a Pearson correlation of $r = 0.503$.
CDT-II advances this proof of concept in three respects. First, it removes the dependence on pre-trained RNA and protein language models, encoding expression directly and enabling cell-level rather than pseudo-bulk prediction. Second, it concentrates on the transcriptional (DNA-to-RNA) stage of the cascade and validates it rigorously on entirely held-out genes. Third, it elevates interpretability---attention maps and gradient-based attribution---from a by-product to the central deliverable.

We validated CDT-II on a large-scale K562 CRISPRi screen \citep{gilbert2013,shalem2014,morris2023}, holding out five perturbation targets entirely---including \textit{GFI1B}, a master transcriptional regulator, and \textit{TFRC} (CD71), the target of the anti-TfR1 antibody PPMX-T003 currently in clinical trials \citep{ogama2023}.

Our results establish three principal claims, developed in turn below. First, CDT-II's attention maps recover known regulatory structure without supervision---a trans-regulatory network, ENCODE regulatory elements including CTCF sites, and an RNA-processing module identified convergently by two independent attention mechanisms. Second, an independent interrogation of the trained model through input gradients recovers the same regulatory relationships and predicts the transcriptional consequences of perturbing therapeutic targets. Third, applied to the held-out drug target \textit{TFRC}, gradient analysis yields a genome-wide regulatory map of TfR1 inhibition that aligns with PPMX-T003 clinical and preclinical findings in five of ten functional categories, while the remaining five---including an ER stress signature not yet reported clinically---are testable predictions awaiting experimental validation.

\section{Methods}

\subsection{Data sources and preprocessing}

\textbf{Primary CRISPRi dataset.} We used the STING-seq v2 dataset \citep{morris2023} (GEO accession GSE171452), a single-cell CRISPRi screen in K562 cells that pairs pooled CRISPR-interference perturbations with single-cell RNA readout. Each perturbation's transcriptome-wide effect is thus measured directly at single-cell resolution---the property CDT-II exploits to learn perturbation responses. It profiled 60,505 K562 cells using single-cell RNA sequencing combined with CRISPR interference \citep{gilbert2013,dixit2016}. Perturbations targeted 447 genomic loci: 27 transcription start sites (TSS) and 420 single-nucleotide polymorphism (SNP) loci. Cells were retained only if exactly one target gene had UMI $\geq 50$ with no competing signal, yielding 10,328 assignable cells (8,250 TSS-perturbed and 2,078 non-targeting controls used as unperturbed baseline). An additional 7,407 SNP-perturbed cells (420 loci) were used with guide assignments from the original publication \citep{morris2023}, for a total of 15,657 perturbed cells (8,250 TSS + 7,407 SNP; 13,620 training, 2,037 validation).

\textbf{Gene set curation.} The 2,361-gene set was derived by intersecting genes detected as expressed in both an independent K562 CRISPRi screen \citep{gasperini2019}---the Gasperini \textit{et al.} enhancer-scale screen, which perturbed thousands of candidate enhancers in K562 and provides a large, independent expression reference---and the primary STING-seq dataset \citep{morris2023}, plus \textit{GFI1B} as a critical test case for unsupervised network recovery. Restricting to genes reproducibly detected across these two independent screens removes assay-specific noise (Results).

\textbf{DNA embeddings.} Pre-computed Enformer \citep{avsec2021enformer} embeddings of shape [896, 3,072] were generated for each perturbation locus. The 896 bins at 128-base-pair resolution span $896 \times 128 \approx 115$ kilobases, i.e. $\pm$57 kb centred on the locus; this window defines the genomic context available to DNA self-attention. We used Enformer because it is an established sequence-to-expression model whose epigenomic pre-training yields embeddings informative for regulatory-element prediction. The DNA embedding component is modular (Discussion): it can be replaced by newer genomic models such as Borzoi \citep{linder2025borzoi}, AlphaGenome \citep{avsec2026alphagen}, or Evo \citep{nguyen2024} without altering the rest of the architecture.

\subsection{Model architecture}

CDT-II follows the design principles established in CDT-I \citep{ota2026cdtv1} (Fig.~1). The RawExpressionEncoder combines learned gene identity embeddings with projected log1p counts-per-million (CPM) expression values to produce gene-level representations. Enformer DNA embeddings are projected from 3,072 to 512 dimensions. All attention operations use scaled dot-product attention \citep{vaswani2017} with 8 heads, embedding dimension $d = 512$, FFN hidden dimension $4d = 2{,}048$, GELU activation, and dropout $p = 0.3$. DNA representations pass through 2 self-attention layers; RNA representations pass through 1 self-attention layer followed by 1 cross-attention layer (RNA queries, DNA keys/values). The Virtual Cell Embedder \citep{ota2026cdtv1} uses 4-head attention pooling to compress each modality to a single vector; these per-modality vectors are then concatenated and fused. A two-layer task head predicts log2 fold changes (log$_2$FC) for all 2,361 genes. Total parameters: $\sim$21 million.

Two design choices follow directly from the central dogma. First, in the cross-attention layer RNA representations form the queries and DNA representations the keys and values, realizing the directional DNA-to-RNA flow of transcriptional control. In attention, the query is the element that seeks information while the keys and values are the source attended to; assigning RNA genes as queries therefore casts each gene as interrogating the genomic context to discover which regulatory positions govern its expression---the biological act of a gene ``reading'' its regulatory landscape. This assignment also fixes the shape of the cross-attention matrix as [genes $\times$ genomic positions] (\emph{Attention map extraction and analysis})---a per-gene distribution over the locus that is directly interpretable as a regulatory-element map, and precisely what the downstream ENCODE/CTCF enrichment analysis operates on. The transposed assignment would instead yield a [positions $\times$ genes] matrix that does not directly answer the biological question of which positions regulate a given gene; recovering a per-gene regulatory map from it would require additional aggregation across positions, complicating the analysis and weakening interpretability. The reverse assignment (DNA querying RNA) would moreover model an RNA-to-DNA relationship that has no counterpart in transcription. Second, the Virtual Cell Embedder receives the DNA representation and the \textit{cross-attention output} (RNA conditioned on DNA), not the pre-cross-attention RNA self-attention map, so that the cell embedding reflects the completed transcriptional-control computation. Passing the intermediate RNA map directly to the embedder would let the model bypass the cross-attention stage and would blur the one-to-one correspondence between each attention mechanism and a specific biological relationship that makes CDT-II interpretable.

\subsection{Training procedure}

The prediction target for each perturbed cell is the per-gene log2 fold change relative to the non-targeting-control (NTC) mean expression, $\log_2((\text{cell}+1)/(\text{NTC mean}+1))$. The NTC baseline is the unperturbed reference and is computed within the dataset; at inference the model receives only the perturbed locus and the cell's raw expression---not the reference profile---and therefore predicts the expression change directly. Because every training target is defined against this fixed cell-line baseline, the model internalizes the unperturbed reference in its parameters; the NTC profile is used only to construct training labels and is never supplied as a model input, at training or inference. This is what enables log2 fold-change prediction without a reference profile. Training used mean squared error loss with AdamW \citep{loshchilov2019} optimizer (learning rate $1 \times 10^{-4}$, weight decay $1 \times 10^{-5}$) and ReduceLROnPlateau scheduling (factor 0.5, patience 10 epochs). Gradient norms were clipped to 1.0. Batch size was 64. Training was performed on a single NVIDIA A100 GPU (40 GB) and completed in approximately 2 days. Inference takes approximately 0.5 seconds per cell. Five genes were held out entirely at the gene level: \textit{GFI1B}, \textit{CD52}, \textit{TFRC}, \textit{CD44}, and \textit{TNFSF9}.

\subsection{Attention map extraction and analysis}\label{subsec:attention}

Attention weights were extracted from each mechanism during inference: the RNA self-attention matrix [$2{,}361 \times 2{,}361$] captures gene co-regulation; the DNA-to-RNA cross-attention matrix [$2{,}361 \times 896$] captures gene-by-position regulatory relationships. Louvain community detection \citep{blondel2008} (resolution 1.0) was applied to identify modules. Gene Ontology \citep{ashburner2000} and pathway enrichment was performed using Enrichr \citep{kuleshov2016} against GO Molecular Function, KEGG \citep{kanehisa2000}, and Reactome \citep{jassal2020} databases.

\subsection{Gradient analysis}

CDT-II receives two inputs: DNA embeddings (Enformer, fixed per target gene) and RNA expression values (2,361 genes per cell). Gradient analysis queries the model's learned regulatory structure by computing the Jacobian matrix $J_{ji} = \partial(\text{predicted output}_j)/\partial(\text{RNA input}_i)$ via backpropagation through the full model---including RNA self-attention, DNA-to-RNA cross-attention, the Virtual Cell Embedder, and output layers. Unlike attention maps, which reflect a single layer, the Jacobian captures the integrated sensitivity across all model components.

For each held-out perturbation target, the DNA embedding was fixed to the target gene's Enformer representation, and RNA input was set to the non-targeting control (NTC) mean expression across 2,361 genes. The Jacobian was computed for the top 100 output genes ranked by experimental effect size (mean $|\log_2\text{FC}|$ across perturbed cells). For each input gene $i$, a gradient importance score was defined as the mean absolute Jacobian across all 100 output genes: $\text{GI}(i) = \frac{1}{100}\sum_{j=1}^{100}|J_{ji}|$. Genes were ranked by this score to identify the input genes whose expression most strongly influences the model's predictions under a given perturbation. Gradient-derived rankings were compared with experimental effect sizes for each of the five held-out genes, yielding mean Pearson $r = 0.82$ (range 0.76--0.86). Importantly, this correlation is not built into the training objective. The model is trained only to predict each gene's output change (log$_2$FC); it is never supervised on which input genes are regulatorily influential. GI($i$) measures gene $i$'s influence \textit{as an input} on the model's predictions, whereas the experimental effect size measures how much gene $i$ \textit{itself} changes \textit{as an output}---distinct quantities, since a gene can change strongly without being influential, or be influential without changing much. That the two nonetheless agree indicates the model has learned that strongly perturbed genes also tend to act as regulatory hubs, an emergent biological signal rather than a tautological relationship.

For the Jacobian heatmap visualization (Fig.~9), the absolute Jacobian values for the top 100 output genes $\times$ top 50 input genes (by gradient importance) were log$_{10}$-transformed and hierarchically clustered using Ward's method with correlation distance.

\subsection{ENCODE enrichment analysis}

\begin{sloppypar}
Cross-attention profiles were compared against ENCODE K562 regulatory
annotations \citep{encode2012}: CTCF ChIP-seq (ENCFF769AUF), H3K27ac ChIP-seq
(ENCFF864OSZ), H3K4me1 ChIP-seq (ENCFF135ZLM), H3K4me3 ChIP-seq
(ENCFF313FYW), and DNase-seq (ENCFF422QRZ).
For each gene, the top 10\% of attention bins (89 of 896) were tested
for overlap with regulatory peaks using Fisher's exact test with Haldane
correction (adding 0.5 to each cell of the $2 \times 2$ table).
All significant $P$ values were below the Bonferroni-corrected threshold
($0.05/25 = 0.002$). Circular permutation tests ($n = 1{,}000$)
controlled for spatial autocorrelation.
\end{sloppypar}

\subsection{Cross-attention CTCF enrichment across 28 CRISPRi target genes}

To test generalizability beyond the five held-out genes, cross-attention was extracted for all 28 CRISPRi target genes---the 27 TSS-perturbed targets plus \textit{GFI1B}---using the Morris STING-seq v2 dataset \citep{morris2023}. To confirm robustness to the choice of CTCF peak calls, we used an independent ENCODE K562 CTCF ChIP-seq dataset (ENCFF796WRU) rather than the dataset used in the 5-gene analysis (ENCFF769AUF). For each gene, the fraction of CTCF sites in the top 10\% attention bins was compared to the expected 10\% baseline. Significance was assessed by permutation test ($n = 1{,}000$): for each permutation, random sets of 89 bins (top 10\% of 896) were selected, and the enrichment was recalculated.

\section{Results}

\subsection{CDT-II architecture and the AI microscope framework}

CDT-II implements a two-modality architecture that follows the directional logic of the central dogma (Fig.~1). Genomic DNA embeddings, generated by Enformer \citep{avsec2021enformer} from sequences centered on perturbation sites, are represented as a $896 \times 3{,}072$ matrix. Per-cell RNA expression values for 2,361 genes are encoded by a RawExpressionEncoder that combines learned gene identity embeddings with projected expression values, producing gene-level representations of dimension 512.

Each modality first passes through self-attention layers that capture internal structure: two layers for DNA and one layer for RNA. A cross-attention layer then models transcriptional regulation, with RNA representations serving as queries and DNA representations as keys and values. DNA and RNA are processed as separate streams throughout and interact only at this cross-attention layer; they are never concatenated into a single input sequence. All attention operations use 8 heads and preserve the original dimensionality, ensuring that attention maps remain directly interpretable as gene-by-gene or gene-by-position matrices.

Because of this design, CDT-II produces two coupled outputs rather than a prediction alone: the quantitative held-out-gene perturbation response (log2 fold changes) and an interpretable gene regulatory network read directly off its intra-RNA, DNA--RNA cross-attention, and gradient maps. The latter is the model's primary deliverable---perturbation-prediction methods that emit expression values alone do not expose which regulatory elements act on which genes in a given cellular state---and it is the basis of CDT-II's novelty (Positioning relative to existing methods; Table~\ref{tab:positioning}).

We validated CDT-II on a large-scale K562 CRISPRi screen \citep{morris2023} comprising 15,657 cells (13,620 training; 2,037 validation). Five perturbation targets were held out entirely at the gene level: \textit{GFI1B}, \textit{CD52}, \textit{TFRC}, \textit{CD44}, and \textit{TNFSF9}.

\begin{figure}[tp]
\centering
\includegraphics[max width=\textwidth,max totalheight=0.82\textheight]{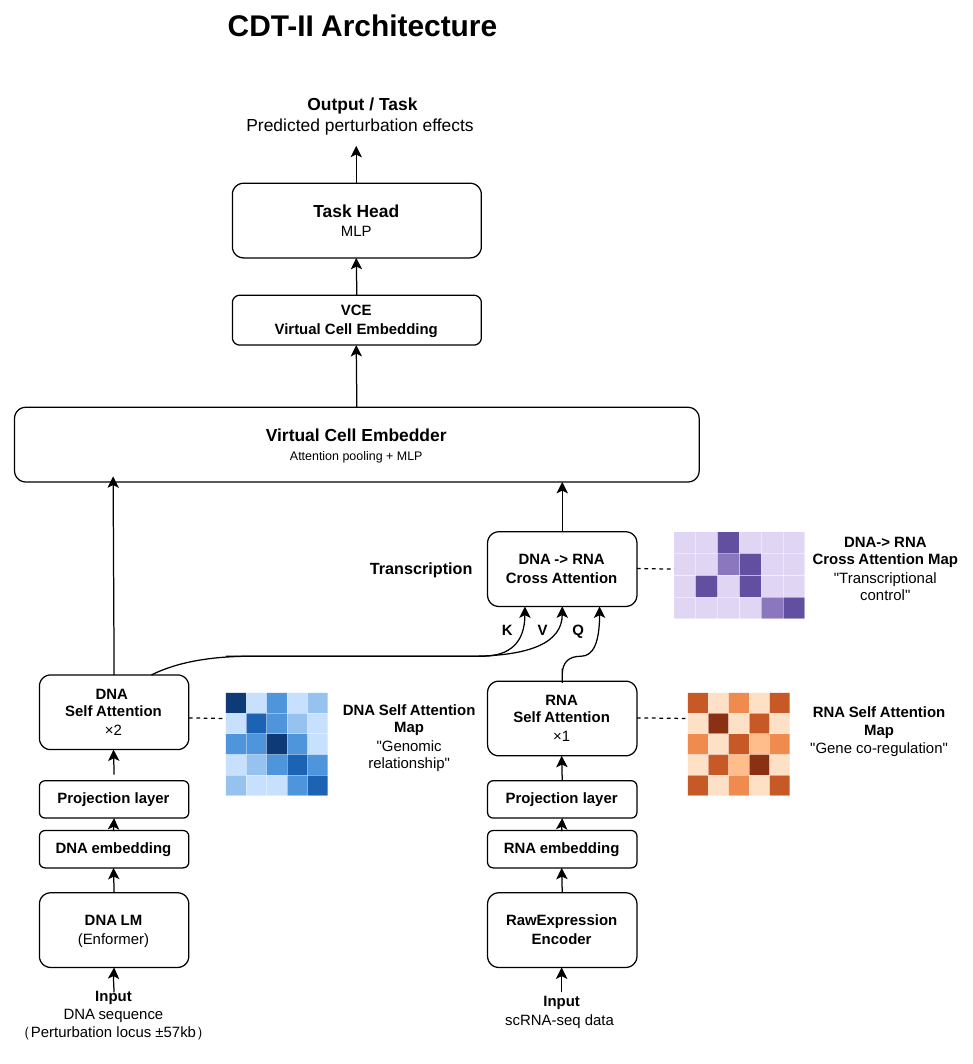}
\caption{\textbf{CDT-II architecture.} The model mirrors the central dogma: DNA self-attention (2 layers, shown as a single block with ``$\times 2$'') captures genomic relationships within a $\pm$57 kb window, RNA self-attention (1 layer) captures gene co-regulation, and cross-attention models transcriptional control. A Virtual Cell Embedder integrates both modalities to predict perturbation effects.}
\label{fig:architecture}
\end{figure}

\subsection{Gene set quality determines model resolution}

Our initial attempt used a larger 9,335-gene set selected by an expression threshold in the primary STING-seq dataset alone (without cross-dataset filtering; $\sim$54 million parameters), plateauing at $r = 0.37$ with attention collapse---that is, attention matrices became near-uniform and diffuse, failing to form the structured gene modules seen with the filtered set. Despite having 2.6$\times$ more parameters, this larger model underperformed, indicating that model capacity is not the bottleneck---data quality is. We hypothesized that genes lacking reproducibility across independent experiments introduced noise. Applying a cross-dataset filter---retaining 2,361 genes detected in two independent K562 CRISPRi screens \citep{morris2023,gasperini2019}---the smaller model ($\sim$21 million parameters) improved steadily to $r = 0.64$ with training $r = 0.65$, indicating minimal overfitting (Fig.~2).

\begin{figure}[tp]
\centering
\includegraphics[max width=\textwidth,max totalheight=0.82\textheight]{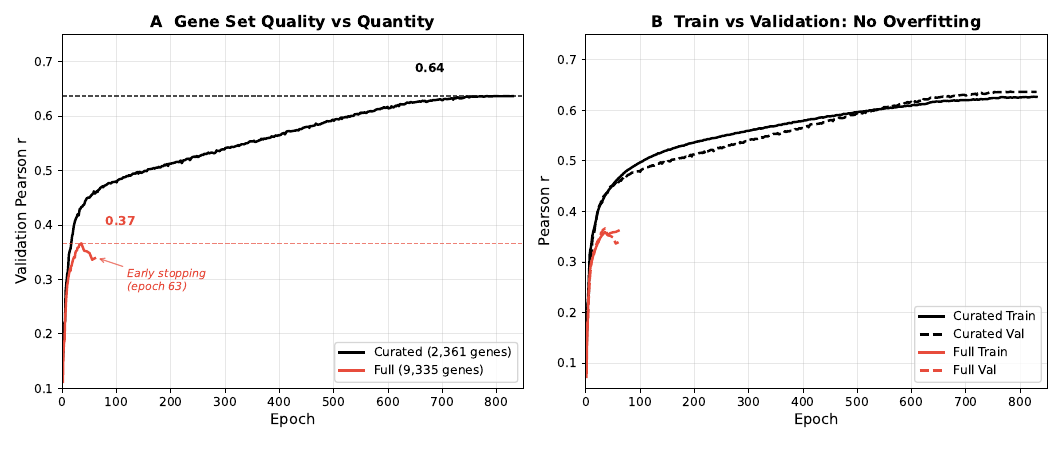}
\caption{\textbf{Gene set quality determines model resolution.} \textbf{(A)} Validation learning curves for the two gene sets: the 9,335-gene model plateaus at $r = 0.37$; the 2,361-gene cross-dataset-filtered model reaches $r = 0.64$. \textbf{(B)} Training and validation curves for both models, showing minimal overfitting.}
\label{fig:learning}
\end{figure}

\subsection{CDT-II predicts cell-level perturbation effects}

CDT-II achieved an overall validation Pearson $r$ of 0.64 ($R^2 = 0.41$),
predicting cell-level log2 fold changes across all five held-out genes simultaneously.
To evaluate per-gene performance, we computed pseudo-bulk correlations on mean
trans-effect profiles: \textit{GFI1B} ($r = 0.88$), \textit{TNFSF9} ($r = 0.86$),
\textit{TFRC} ($r = 0.85$), \textit{CD44} ($r = 0.85$), and \textit{CD52}
($r = 0.75$), yielding a mean of $r = 0.84$ (Fig.~3).
These correlations should be read against the reproducibility of the underlying
measurements: Morris et al.\ found that only 56\% of \textit{GFI1B} trans-targets
replicated in the independent Gasperini dataset \citep{morris2023,gasperini2019}, and direct comparison of
pseudo-bulk effect sizes between the two studies yields $r \approx 0.79$.
Single-cell CRISPRi readouts are thus intrinsically noisy, which bounds the
perturbation-specific signal recoverable against a single experimental replicate
and is reflected in the cell-level correlation ($r = 0.64$). We therefore interpret
CDT-II's correlations as recovery of reproducible biological signal within K562
rather than as a performance score (see \emph{Positioning relative to existing methods}).

\begin{figure}[tp]
\centering
\includegraphics[max width=\textwidth,max totalheight=0.82\textheight]{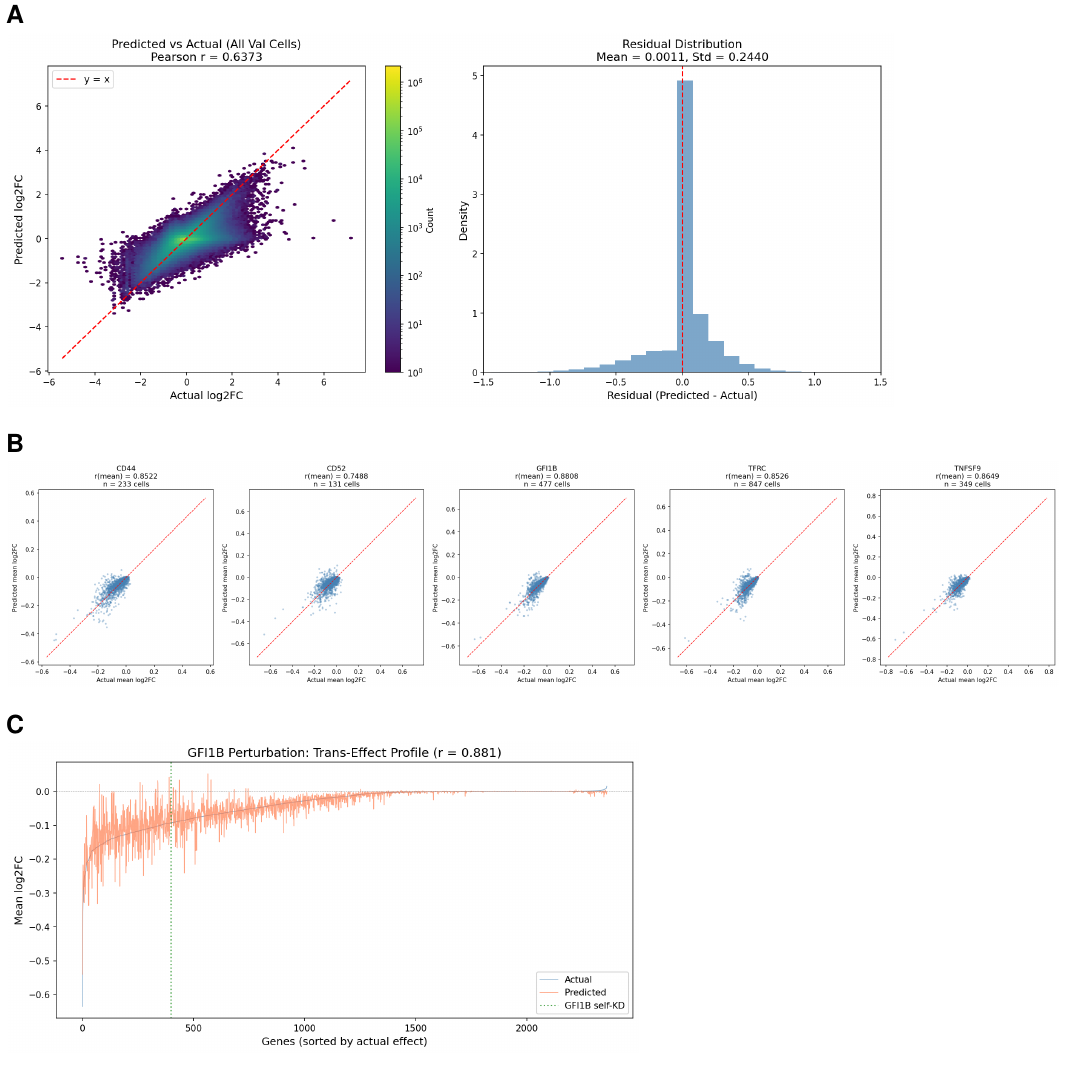}
\caption{\textbf{CDT-II predicts cell-level perturbation effects.} \textbf{(A)} Left: density scatter of predicted versus observed cell-level log2 fold changes across all validation cells (Pearson $r = 0.64$; dashed line $y=x$). Right: distribution of residuals (predicted $-$ observed), centred near zero (mean $= 0.001$, s.d.\ $= 0.24$), indicating that predictions are essentially unbiased with small errors. \textbf{(B)} Per-gene pseudo-bulk correlations for five held-out genes (mean $r = 0.84$). \textbf{(C)} Trans-effect prediction for \textit{GFI1B}.}
\label{fig:predictions}
\end{figure}

\subsection{Attention maps reveal the \textit{GFI1B} regulatory network}

The RNA self-attention matrix captures learned co-regulatory relationships. Extracting the \textit{GFI1B} query row revealed biologically coherent rankings: cell cycle regulators including \textit{CDCA8}, \textit{CDC20}, \textit{KIF2C}, and \textit{KIF14} appeared in the top ranks, consistent with \textit{GFI1B}'s role in cell cycle control during hematopoietic differentiation \citep{saleque2007}.

Ranking all 2,361 genes by \textit{GFI1B} attention weight and, independently, by absolute experimental effect size, we measured overlap between the top $N$ genes. At top 100, 28 genes appeared in both lists, representing 6.6-fold enrichment ($P = 3.5 \times 10^{-17}$ by hypergeometric test). This demonstrates that CDT-II's attention maps recover known regulatory relationships without supervision (Figs.~4 and 5).

\begin{figure}[tp]
\centering
\includegraphics[max width=\textwidth,max totalheight=0.82\textheight]{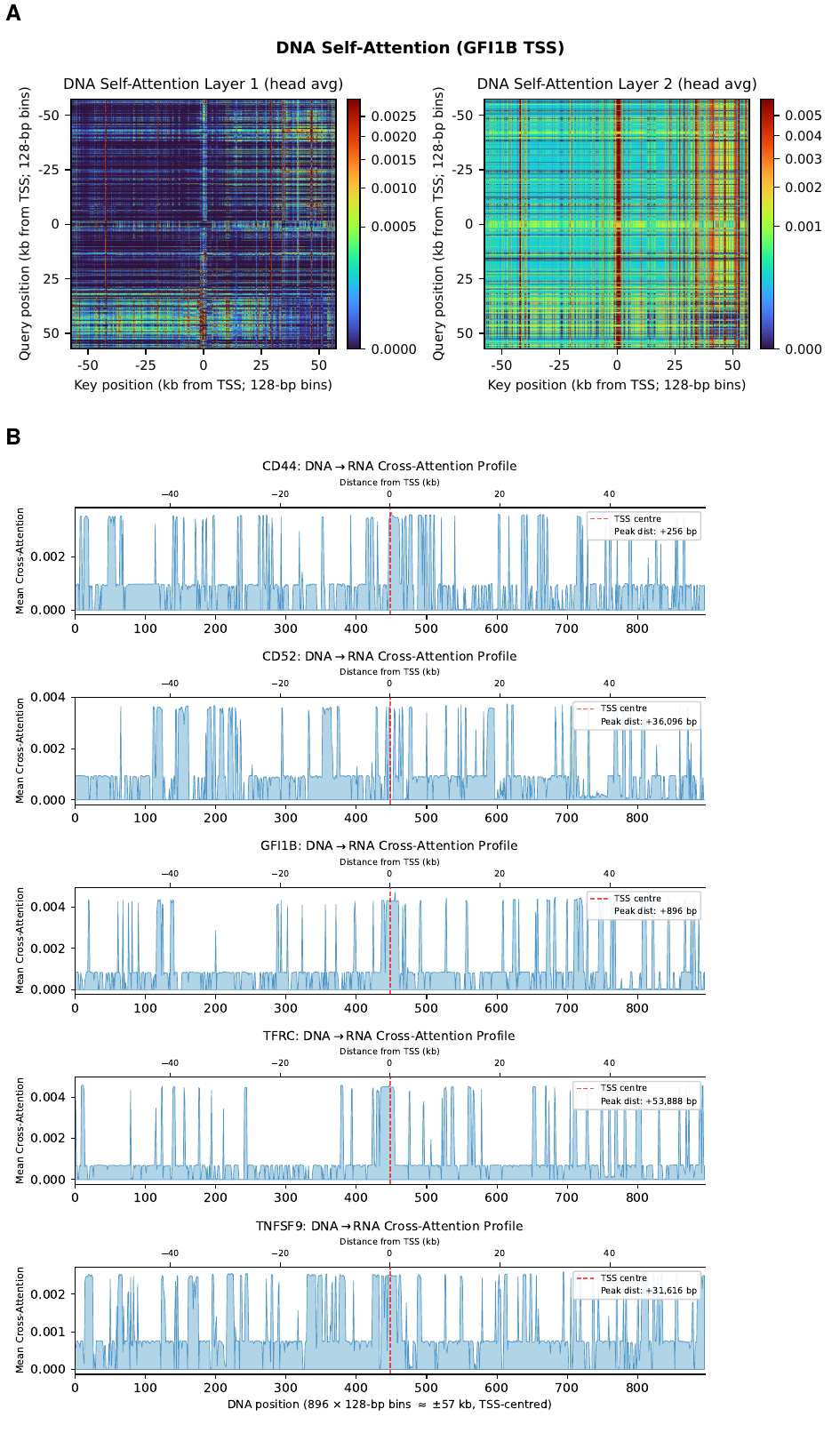}
\caption{\textbf{DNA and cross-attention maps reveal regulatory structure.} \textbf{(A)} DNA self-attention (head-averaged) for a held-out gene (\textit{GFI1B} TSS). Both axes are the 896 Enformer genomic bins (128 bp each, spanning $\pm$57 kb around the TSS); colour is the attention weight from each query bin to each key bin. Bright vertical bands mark genomic positions that broadly influence others, including the TSS-centred bin. \textbf{(B)} DNA-to-RNA cross-attention profiles: one trace per held-out gene, showing the mean cross-attention weight across the 896 genomic bins, with the TSS-centred bin marked. Higher weight indicates a stronger learned DNA$\to$RNA regulatory link.}
\label{fig:attention_dna}
\end{figure}

\begin{figure}[tp]
\centering
\includegraphics[max width=\textwidth,max totalheight=0.82\textheight]{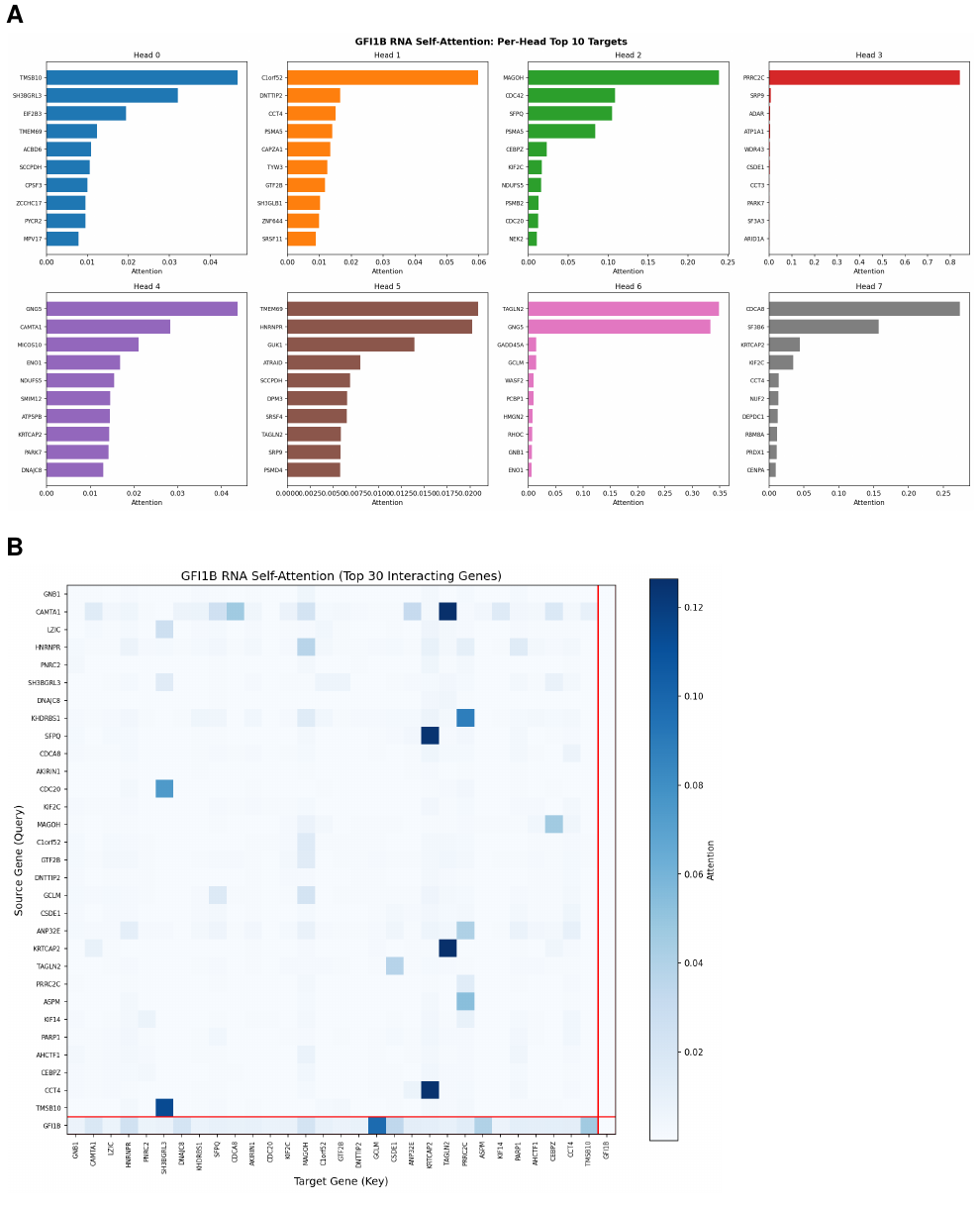}
\caption{\textbf{RNA self-attention maps reveal gene co-regulation.} \textbf{(A)} Per-head attention patterns, showing distinct co-regulatory programs learned by each head. \textbf{(B)} Gene-by-gene heatmap of RNA self-attention among the 30 genes interacting most strongly with \textit{GFI1B}.}
\label{fig:attention2}
\end{figure}

\subsection{Convergent attention reveals an RNA processing module}

We applied Louvain community detection \citep{blondel2008} to networks derived from each mechanism separately. From the RNA self-attention matrix, community C1 (679 genes) showed enrichment for RNA binding (Gene Ontology \citep{ashburner2000}; $P = 5 \times 10^{-16}$) and metabolism of RNA (Reactome \citep{jassal2020}; $P = 2 \times 10^{-9}$). From the cross-attention matrix, community C3 (165 genes) showed even stronger enrichment for RNA binding ($P = 1 \times 10^{-16}$), spliceosome (KEGG \citep{kanehisa2000}; $P = 4 \times 10^{-10}$), and mRNA splicing (Reactome \citep{jassal2020}; $P = 2 \times 10^{-9}$). Community assignments and full gene lists for all communities from both mechanisms (Supplementary Data~S1), and the corresponding GO and pathway enrichment results, are provided in the code repository.

The key finding is convergence: 132 of 165 cross-attention genes (80\%) were contained within the 679-gene self-attention community (2.8-fold enrichment over the 47.5 expected by chance; hypergeometric $P = 9.3 \times 10^{-46}$), despite being identified through fundamentally different mechanisms---one operating on gene-gene relationships, the other on gene-position relationships (Fig.~6). Such convergence is meaningful precisely because the two mechanisms take different inputs and optimize different relationships: their independent agreement on the same gene set is evidence that the module reflects genuine biological co-regulation rather than an artifact of any single mechanism, analogous to concordant readouts from two independent assays.

\begin{figure}[tp]
\centering
\includegraphics[max width=\textwidth,max totalheight=0.82\textheight]{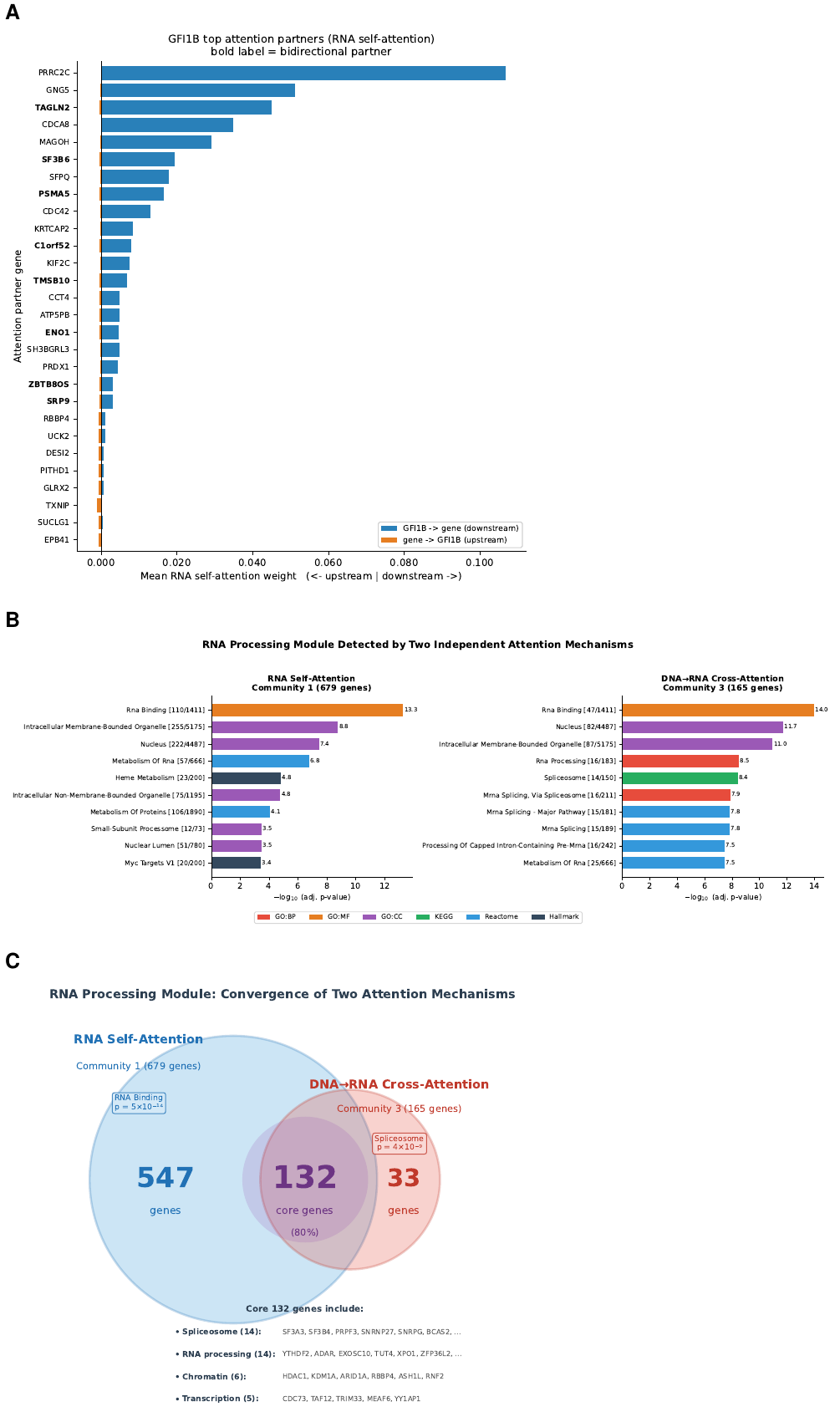}
\caption{\textbf{Convergent attention reveals an RNA processing module.} \textbf{(A)} \textit{GFI1B}'s top RNA self-attention partners, ranked by attention weight and split by direction---downstream (\textit{GFI1B}$\rightarrow$gene) versus upstream (gene$\rightarrow$\textit{GFI1B}); bold labels mark bidirectional partners. \textbf{(B)} GO enrichment comparison between RNA self-attention and cross-attention communities. \textbf{(C)} Venn diagram of the RNA self-attention community (679 genes: 547 unique $+$ 132 shared) and the DNA--RNA cross-attention community (165 genes: 33 unique $+$ 132 shared). The 132-gene core---representing 80\% of the cross-attention community (hypergeometric $P = 9.3 \times 10^{-46}$)---is listed below the diagram, grouped by functional category (spliceosome, RNA processing, chromatin, and transcription), showing that the module recovered by both mechanisms is dominated by RNA-processing machinery.}
\label{fig:convergent}
\end{figure}

\subsection{Cross-attention recovers known regulatory elements}

To test whether cross-attention learns genuine regulatory structure, we compared attention patterns against ENCODE regulatory annotations \citep{encode2012} for K562 cells (Fig.~7).

Enformer \citep{avsec2021enformer}, which provides CDT-II's DNA embeddings, was pre-trained on epigenomic tracks, so the initial embeddings already encode regulatory element positions. However, cross-attention does not operate on these raw embeddings: a projection layer and two DNA self-attention layers, trained end-to-end on perturbation prediction, substantially transform the representations before cross-attention.

CDT-II's cross-attention showed strong correspondence with ENCODE regulatory elements. Of 25 gene--mark combinations, 24 were testable (CD44 $\times$ H3K27ac yielded no ENCODE peaks in the window); of these, 23 achieved Fisher's exact $P < 0.001$. The strongest enrichments were observed for DNase hypersensitive sites (\textit{GFI1B}: 201$\times$ odds ratio) and CTCF binding (\textit{CD52}: 28$\times$). H3K27ac enrichment was consistently high across four testable genes (range 4--11$\times$), confirming that cross-attention preferentially targets active regulatory regions \citep{fulco2019}. Circular permutation tests ($n = 1{,}000$) confirmed that these enrichments are robust to spatial autocorrelation, and results remained significant across attention thresholds from top 5\% to top 20\%.

\begin{figure}[tp]
\centering
\includegraphics[max width=\textwidth,max totalheight=0.82\textheight]{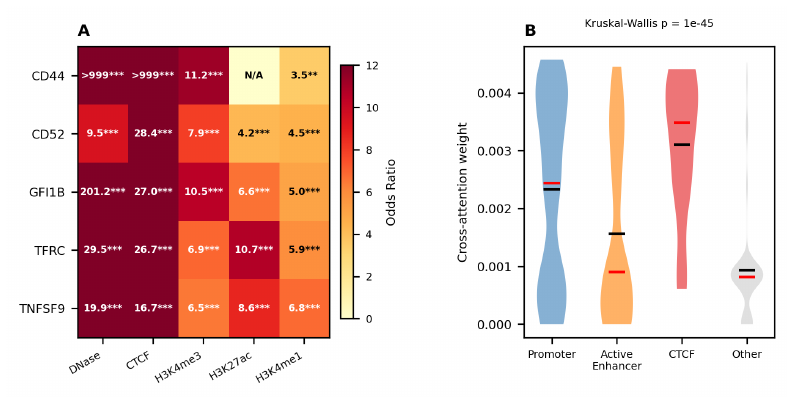}
\caption{\textbf{Cross-attention recovers known regulatory elements.} \textbf{(A)} Fisher's exact test enrichment for five ENCODE K562 marks across five held-out genes. 23/24 combinations reach $P < 0.001$. \textbf{(B)} Distribution of cross-attention weight across ENCODE bin classes (promoter, active enhancer, CTCF, other), pooled over the five held-out genes; black and red bars mark the mean and median, respectively (Kruskal--Wallis $P = 1 \times 10^{-45}$).}
\label{fig:encode}
\end{figure}

\subsection{CTCF enrichment generalizes across 28 CRISPRi target genes}

To test whether cross-attention CTCF enrichment generalizes beyond the five held-out genes, we extracted cross-attention weights for all 28 CRISPRi target genes in the Morris dataset \citep{morris2023} and overlaid an independent ENCODE K562 CTCF ChIP-seq dataset (ENCFF796WRU), distinct from the one used in the 5-gene ENCODE analysis above.

Cross-attention showed strong CTCF enrichment across essentially all genes: mean 7.67$\times$ (median 8.57$\times$), with 26/28 genes exceeding 2$\times$ and 21/28 exceeding 5$\times$ enrichment. Permutation testing ($n = 1{,}000$) confirmed statistical significance: observed 7.67$\times$ versus null $1.00 \pm 0.27$ ($P < 0.001$). Notable examples include TFRC (10.0$\times$), ITGB1 (10.0$\times$), ENTPD1 (10.0$\times$), and CD52 (8.6$\times$). The consistency of CTCF enrichment across both the original 5-gene analysis and this independent 28-gene validation with a different peak-call dataset confirms the robustness of this finding.

This finding is notable because the model receives no explicit chromatin conformation data. CTCF is a primary architectural protein establishing topologically associating domain (TAD) boundaries and chromatin loop anchors \citep{dixon2012}. The consistent enrichment across 28 independent loci suggests that CDT-II learns aspects of three-dimensional chromatin organization from sequence-derived genomic representations and expression data alone.

\subsection{Gradient-based attribution recovers regulatory relationships}

The preceding analyses demonstrate that CDT-II's attention maps capture biologically meaningful structure. However, attention weights reflect learned representations that are not straightforward to interpret as quantitative predictions of regulatory influence \citep{jain2019,wiegreffe2019}. We therefore asked whether an independent interrogation method---input gradients---could extract quantitative regulatory relationships from the trained model.

As a complementary approach to attention maps, we computed the Jacobian matrix $J_{ji} = \partial(\text{predicted output}_j)/\partial(\text{RNA input}_i)$ for each held-out perturbation target (see Methods). The Jacobian captures the integrated sensitivity of the model's output to each RNA input gene across all layers, rather than the learned representation at any single layer. For each input gene $i$, a gradient importance score---the mean $|J_{ji}|$ across the top 100 experimentally affected output genes, $\text{GI}(i) = \frac{1}{100}\sum_{j=1}^{100} |J_{ji}|$ (Methods)---quantifies how strongly the model relies on that gene's expression to predict downstream perturbation effects.

We compared these gradient importance scores with experimental CRISPRi effect sizes (mean $|\log_2\text{FC}|$ across perturbed cells). Across all 2,361 genes, gradient-derived importance and experimental effect sizes showed strong agreement: mean Pearson $r = 0.82$ across five held-out genes (range 0.76--0.86; Fig.~8A). The top 20 genes by gradient importance (orange points in Fig.~8A) cluster along the regression line, indicating that the genes the model considers most influential are also those with the largest experimental effects. This correlation was consistent across genes with different numbers of perturbed cells and different effect magnitudes (Fig.~8B).

The five held-out genes are themselves established or emerging therapeutic targets: \textit{TFRC} (CD71) is an antibody--drug conjugate (ADC) target in acute leukemia, \textit{CD52} is the target of alemtuzumab in chronic lymphocytic leukemia and multiple sclerosis, \textit{CD44} is targeted by multiple antibodies in clinical development for acute myeloid leukemia, \textit{TNFSF9} (4-1BB ligand) is an immuno-oncology target, and \textit{GFI1B} is a hematopoietic transcription factor implicated in leukemogenesis. That gradient-based attribution predicts the downstream transcriptional consequences of perturbing each of these targets ($r = 0.76$--$0.86$) has direct translational relevance: the gradient importance profile for a drug target gene provides a computational prediction of which pathways will be affected by therapeutic intervention, enabling virtual screening for on-target efficacy and off-target effects before committing to costly experimental validation.

This result has three important implications. First, it provides independent validation that CDT-II has learned genuine regulatory relationships---the gradient method queries the model through a completely different mechanism than attention maps, yet recovers the same biological structure. Second, it supports the use of gradient-based attribution as a practical tool for virtual regulatory screening: for any gene with Enformer embeddings, the gradient importance profile predicts which downstream genes will be most affected by perturbation, without requiring new experiments. Third, for therapeutic targets, gradient profiles offer a principled approach to anticipating the transcriptional consequences of drug action---identifying not only which genes respond most strongly (potential efficacy biomarkers) but also which unexpected pathways are affected (potential side effects or opportunities for combination therapy).

Notably, conventional in silico knockdown---setting a gene's expression to zero and observing predicted changes---yielded much weaker correlations ($r \approx 0.07$). This contrast highlights a critical distinction for drug target assessment: in silico knockdown simulates complete target ablation, an extreme state far outside the training distribution that produces unreliable predictions. Gradient analysis instead queries the model locally within the training distribution, faithfully recovering the regulatory relationships the model has learned. In therapeutic terms, this mirrors the difference between total gene knockout and the partial modulation achieved by most drugs.

\subsection{TFRC case study: predicting anti-TfR1 antibody effects}

To illustrate the translational potential of gradient analysis, we examined the Jacobian regulatory map for \textit{TFRC} (CD71, transferrin receptor 1), the target of PPMX-T003, an anti-TfR1 monoclonal antibody currently in Phase~1 clinical development (a first-in-human study in healthy volunteers \citep{ogama2023}); the antibody is being developed for polycythemia vera. Because TFRC was held out entirely during training, the model has never observed any TFRC perturbation data---the following regulatory map is a purely computational prediction. The Jacobian heatmap (Fig.~9) visualizes the regulatory relationships under TFRC perturbation: rows represent the top 100 output genes most affected by TFRC knockdown in the CRISPRi experiment, columns represent the top 50 input genes whose expression most strongly influences the model's predictions, and color intensity indicates the strength of regulatory coupling between each gene pair.

Table~1 categorizes all 30 top-ranking input genes by functional pathway. Of the 30 genes, 24 can be linked to known consequences of iron depletion or TfR1 inhibition: oxidative stress and ferroptosis (4 genes), iron-dependent DNA synthesis and repair (3), erythrocyte structure and cytoskeleton (2), immune surface molecules (3), mitochondrial metabolism (2), ER stress and protein homeostasis (5), hematopoietic differentiation (1), vesicle trafficking (1), cell survival signaling (1), and RNA processing (2). The remaining 6 genes involve general transcription and chromatin functions, consistent with the broad growth arrest expected from iron deprivation.

Several of these categories correspond directly to PPMX-T003's reported findings: the erythrocyte/cytoskeleton genes (\textit{EPB41}, \textit{ACTR2}) are consistent with the anemia observed in the Phase~1 trial (sustained decreases in hemoglobin and hematocrit from day~7) \citep{ogama2023}; the iron-dependent DNA synthesis genes (\textit{RRM2}, \textit{RPA2}, \textit{UBE2T}) explain the $>$50\% reticulocyte decrease by day~3 \citep{ogama2023}; the oxidative stress cluster (\textit{GCLM}, \textit{MGST3}, \textit{PGD}, \textit{SH3BGRL3}) aligns with the ferroptosis mechanism demonstrated in preclinical studies \citep{fauzi2025}; the immune surface molecules (\textit{CD46}, \textit{PTPRC}, \textit{CD53}) are potentially relevant to the infusion-related reactions observed in five participants \citep{ogama2023}; and the ER stress genes (\textit{PDIA6}, \textit{SSR2}, \textit{TMCO1}, \textit{RTN4}, \textit{PSMB4}) point to proteostatic consequences of iron depletion not yet characterized clinically.

These associations emerge entirely from CDT-II's learned regulatory structure without any clinical data as input. The gradient analysis produces a comprehensive regulatory map across all 2,361 genes for any perturbation target---the clinical correspondences highlighted here represent a small fraction of the full output. Five of the ten functional categories in Table~1 correspond directly to PPMX-T003 clinical or preclinical findings---infusion-related reactions, ferroptosis, reticulocyte decrease, anemia, and iron-dependent metabolic disruption---serving as a calibration check indicating that the model has learned biologically faithful regulatory structure. The remaining categories, including an ER stress signature (5 genes) and hematopoietic differentiation, represent predictions that extend beyond current clinical characterization of TfR1 inhibition.

\begin{table}[H]
\centering
\footnotesize
\caption{\textbf{Top 30 genes by gradient importance under TFRC perturbation, grouped by functional category.} Genes are ranked by gradient importance (GI) score = mean $|J_{ji}|$ across the top 100 output genes. Categories reflect known roles in iron biology and cellular responses to iron depletion. Clinical relevance indicates correspondence with PPMX-T003 Phase~1 findings \citep{ogama2023} and preclinical studies \citep{fauzi2025}.}
\label{tab:tfrc_top30}
\begin{tabular}{clrll}
\toprule
\textbf{Rank} & \textbf{Gene} & \textbf{GI} & \textbf{Functional category} & \textbf{Clinical relevance} \\
\midrule
1 & \textit{CD46} & 0.0368 & Immune surface molecule & Infusion reactions \\
8 & \textit{CD53} & 0.0333 & Immune surface molecule & Infusion reactions \\
10 & \textit{PTPRC} & 0.0323 & Immune surface molecule & Infusion reactions \\
\midrule
6 & \textit{MGST3} & 0.0336 & Oxidative stress / ferroptosis & Ferroptosis (preclinical) \\
13 & \textit{GCLM} & 0.0313 & Oxidative stress / ferroptosis & Ferroptosis (preclinical) \\
21 & \textit{PGD} & 0.0298 & Oxidative stress / ferroptosis & Ferroptosis (preclinical) \\
29 & \textit{SH3BGRL3} & 0.0289 & Oxidative stress / ferroptosis & Ferroptosis (preclinical) \\
\midrule
26 & \textit{RRM2} & 0.0290 & Iron-dependent DNA synthesis & Reticulocyte decrease \\
25 & \textit{RPA2} & 0.0292 & DNA replication / repair & Reticulocyte decrease \\
27 & \textit{UBE2T} & 0.0290 & DNA repair (Fanconi anemia) & Reticulocyte decrease \\
\midrule
12 & \textit{EPB41} & 0.0313 & Erythrocyte cytoskeleton & Anemia \\
30 & \textit{ACTR2} & 0.0288 & Actin cytoskeleton & Anemia \\
\midrule
3 & \textit{HADHA} & 0.0349 & Mitochondrial metabolism & Iron-dependent \\
9 & \textit{NRDC} & 0.0325 & Metalloprotease & Iron-dependent \\
\midrule
4 & \textit{PDIA6} & 0.0343 & ER stress / protein homeostasis & --- \\
16 & \textit{SSR2} & 0.0308 & ER protein translocation & --- \\
22 & \textit{TMCO1} & 0.0295 & ER calcium homeostasis & --- \\
24 & \textit{PSMB4} & 0.0293 & Proteasome & --- \\
28 & \textit{RTN4} & 0.0290 & ER structure & --- \\
\midrule
5 & \textit{HES4} & 0.0341 & Hematopoietic differentiation & --- \\
\midrule
11 & \textit{RAB10} & 0.0315 & Vesicle trafficking (TfR1 recycling) & --- \\
\midrule
18 & \textit{CCDC88A} & 0.0306 & Cell survival (PI3K-Akt) & --- \\
\midrule
2 & \textit{RSRP1} & 0.0352 & RNA processing & --- \\
19 & \textit{SYF2} & 0.0302 & RNA splicing & --- \\
\midrule
7 & \textit{PITHD1} & 0.0335 & General transcription / chromatin & --- \\
14 & \textit{GNAI3} & 0.0312 & G protein signaling & --- \\
15 & \textit{TAF12} & 0.0311 & Transcription & --- \\
17 & \textit{HMGN2} & 0.0307 & Chromatin structure & --- \\
20 & \textit{PDE4DIP} & 0.0299 & Centrosome & --- \\
23 & \textit{NOC2L} & 0.0294 & Transcriptional repression & --- \\
\bottomrule
\end{tabular}
\end{table}

\begin{figure}[tp]
\centering
\includegraphics[max width=\textwidth,max totalheight=0.82\textheight]{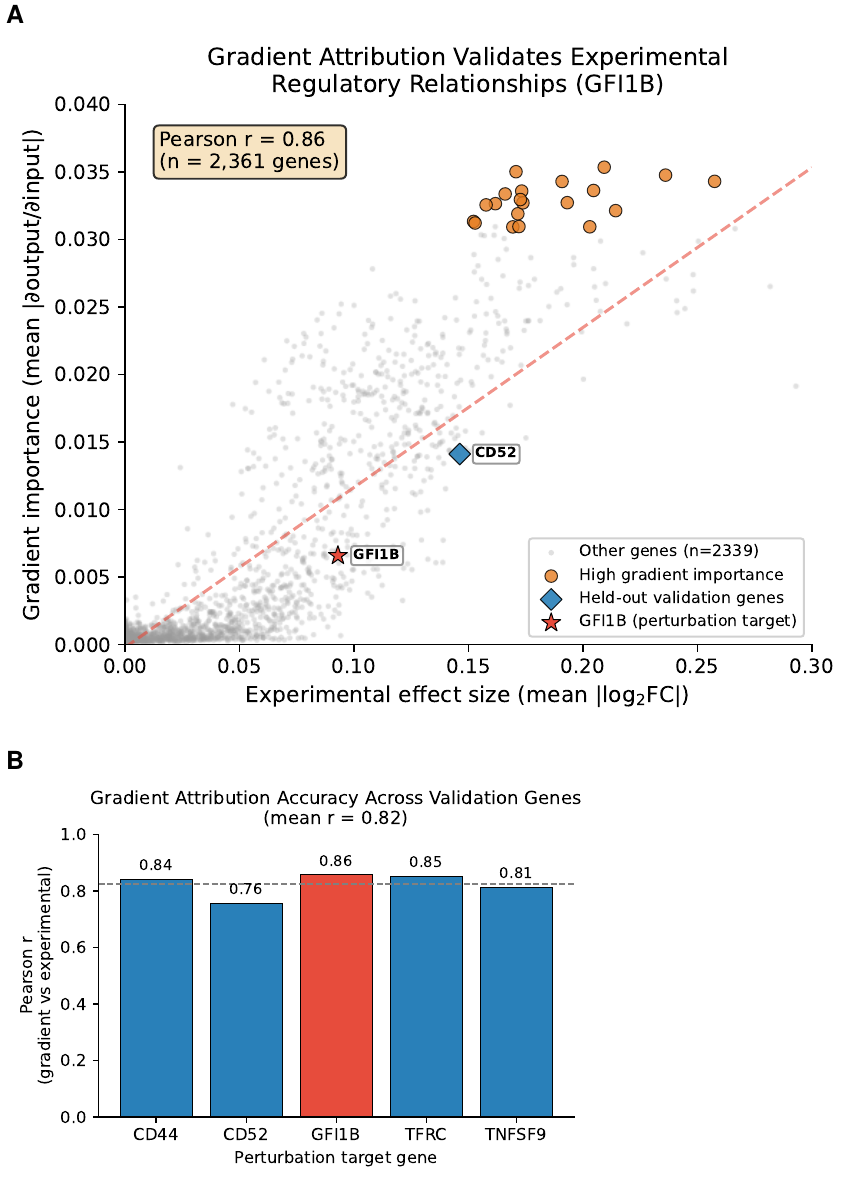}
\caption{\textbf{Gradient-based attribution recovers experimental regulatory relationships.} \textbf{(A)} Scatter plot of gradient importance (mean $|\partial\text{output}/\partial\text{input}|$) versus experimental effect size (mean $|\log_2\text{FC}|$) for 2,361 genes under \textit{GFI1B} perturbation (Pearson $r = 0.86$). Orange points: top 20 genes by gradient importance. Axes are restricted to the region of interest; five genes (0.2\%) with larger effect sizes fall outside the plotted range (the reported Pearson $r$ is computed over all 2,361 genes). Note: CD44, TFRC, and TNFSF9 are among the 28 CRISPRi target genes but fall outside the 2,361-gene expression input set and therefore have no gradient importance scores. \textbf{(B)} Gradient--experimental correlation for each of the five held-out perturbation targets (mean $r = 0.82$); the \textit{GFI1B} bar is highlighted in red because it is the target used as the worked example in panel (A), and the dashed line marks the mean ($r = 0.82$).}
\label{fig:gradient}
\end{figure}

\begin{figure}[tp]
\centering
\includegraphics[max width=\textwidth,max totalheight=0.82\textheight]{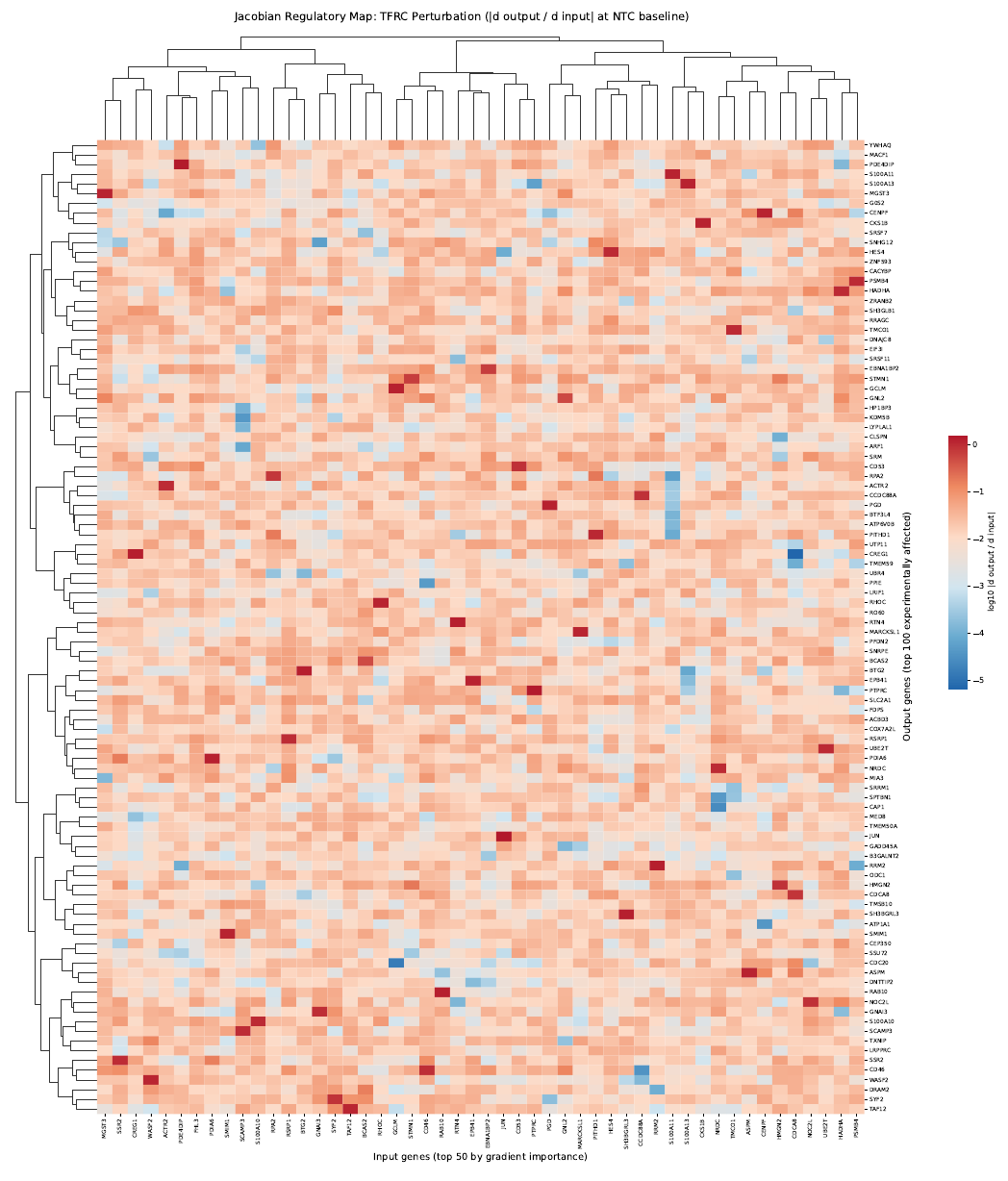}
\caption{\textbf{Jacobian regulatory map for TFRC perturbation.} Clustered heatmap of $|\partial(\text{output gene})/\partial(\text{RNA input gene})|$ under \textit{TFRC} perturbation. Rows: top 100 output genes most affected by TFRC knockdown in the CRISPRi experiment (ranked by experimental $|\log_2\text{FC}|$). Columns: top 50 input genes whose expression most strongly influences the model's predictions (ranked by gradient importance). Color: regulatory influence strength (blue: minimal; red: strong). Row and column dendrograms (Ward's linkage, correlation distance) group co-regulated gene modules.}
\label{fig:jacobian_tfrc}
\end{figure}

\section{Discussion}

CDT-II demonstrates that structuring a neural network to mirror the central dogma produces an ``AI microscope'' whose attention maps recover known regulatory relationships. The key design principle is that each attention mechanism corresponds to a specific biological relationship, making learned representations directly interrogable.

Enformer's pre-training on epigenomic tracks means the initial DNA embeddings encode \textit{general} regulatory element positions based on sequence alone---Enformer receives no cell-type information. These cell-type-agnostic embeddings are then substantially transformed through a learned projection layer and two DNA self-attention layers, all trained on K562 perturbation prediction. The cross-attention maps therefore reflect CDT-II's cell-type-specific understanding rather than Enformer's general-purpose representations. CDT-II is designed as an integration platform: its DNA embedding module is deliberately modular---Enformer can be replaced with newer models such as AlphaGenome \citep{avsec2026alphagen} or Evo \citep{nguyen2024}.

The CTCF enrichment result (mean 7.67$\times$ across 28 CRISPRi target genes, $P < 0.001$) has implications for understanding three-dimensional chromatin organization. That the model, receiving only one-dimensional sequence embeddings and expression data, autonomously highlights CTCF-associated architectural sites suggests emergent learning of chromatin topology. This aligns with the finding that CTCF-associated bins receive the highest attention weights across all five held-out genes (Cohen's $d = 2.4$) in the ENCODE analysis. Future integration of Hi-C contact data could further validate this observation.

The ablation study shows a practical principle: gene set quality, determined by cross-dataset reproducibility filtering, governs model resolution. Training with 9,335 genes produced $r = 0.37$, while 2,361 validated genes yielded $r = 0.64$ with interpretable attention maps. This has direct implications for applying CDT-II to new systems.

CDT-II's architectural simplicity is deliberate: only standard transformer components are used, without specialized modules, custom loss functions, or complex training procedures. This parallels the interpretability-by-design principle demonstrated by BPNet \citep{avsec2021bpnet} for transcription factor binding, extended here to multi-modal regulatory network inference.

Together, the \textit{GFI1B} and \textit{TFRC} analyses illustrate complementary uses of the AI microscope. Attention analysis of \textit{GFI1B} validates CDT-II's ability to recover known regulatory networks---the 6.6-fold enrichment, ENCODE correspondence, and convergent RNA processing module all indicate that these attention-derived maps are biologically faithful. Gradient analysis of \textit{TFRC} then demonstrates translational application: the same instrument, applied to a therapeutic target, generates a genome-wide regulatory map that both aligns with clinical observations and extends beyond them. The two analyses use different interrogation methods (attention versus gradients) on different genes, yet converge on the same conclusion: CDT-II has internalized regulatory structure.

Gradient analysis complements attention maps by capturing the integrated sensitivity across all model layers rather than learned representations at a single layer. That gradient importance correlates with experimental effect sizes (mean $r = 0.82$) independently validates that CDT-II has learned regulatory relationships, and the Jacobian heatmap (Fig.~9) reveals co-regulated modules within these relationships.

\subsection{Positioning relative to existing methods}

CDT-II occupies a methodological niche that is not directly comparable to existing perturbation- or sequence-modelling tools, and we therefore position it by capability rather than by a shared accuracy benchmark. CDT-II's purpose is mechanistic: it is an interpretable instrument whose deliverable is the regulatory structure that its attention maps and gradient attributions expose, with predictive accuracy serving as evidence that this structure is faithful rather than as an end in itself. A task-accuracy leaderboard would therefore evaluate a different objective from the one the model is built to serve. Table~\ref{tab:positioning} contrasts CDT-II with four representative method classes along the dimensions that determine whether a head-to-head comparison is well-posed: the input modalities a method consumes, the output it produces, whether predictions are cell-state-specific, whether they are grounded in measured perturbation data, and the mechanism by which the method is interpretable.

Three distinctions follow. First, sequence-to-function models such as Enformer \citep{avsec2021enformer} and central-dogma \textit{sequence} models operate on molecular sequence and are cell-type-agnostic: they do not consume per-cell expression and do not predict cell-specific responses to a perturbation. Second, gene regulatory network inference methods (SCENIC \citep{aibar2017}, CellOracle \citep{kamimoto2023}) and perturbation-response predictors (GEARS \citep{roohani2024,norman2019}, CPA \citep{lotfollahi2023}) consume single-cell expression but either output a static network or take only expression and a perturbation label as input, without the genomic-sequence context that CDT-II integrates through DNA--RNA cross-attention. Third, CDT-II is, to our knowledge, the only method that jointly consumes genomic-sequence context and the per-cell transcriptome and predicts the measured response of held-out genes to CRISPRi perturbation, while exposing the learned regulatory structure at each layer---genome regulation through intra-DNA self-attention, gene regulation through intra-RNA self-attention, and transcriptional regulation through DNA--RNA cross-attention---together with gradient-based attribution. Because no existing method shares both CDT-II's inputs and its output, a conventional accuracy leaderboard would compare models solving different problems. This absence of a directly comparable method is itself the substance of CDT-II's novelty, which lies in the combination of inputs, output, and mechanistic objective rather than in an incremental gain on an established task.

This combination is not a narrow niche but the minimal one required to read regulatory mechanism off a cell. Only by consuming both genomic context and the cell's own transcriptome, and by predicting the measured response of unseen genes, can a model expose which regulatory elements act on which genes \emph{in that cellular state} and have that structure checked at the bench---a capability none of the above method classes provides. A nascent line of work does begin to combine these two input types, injecting DNA-sequence-model embeddings into a single-cell foundation model (e.g.\ dynamic token adaptation \citep{zhao2025dynamic}) to bring genomic context into single-cell representations. That work, however, augments a representation model and probes regulation by \emph{in silico} sequence mutagenesis, rather than predicting the measured CRISPRi response of held-out genes or casting the directional DNA-to-RNA flow of the central dogma into the architecture so that the trained model yields a directly readable gene regulatory network.

The architectural element of CDT-II most closely related to prior work is the use of cross-attention over genomic features. EPBDxDNABERT-2 \citep{kabir2024epbd} and iEnhancer-Fusion \citep{luo2025ienhancer} both combine the DNABERT-2 language model with DNA-breathing biophysical features through cross-attention, for transcription-factor binding and enhancer identification, respectively; they exemplify a growing body of work that augments sequence language models with biophysical or structural features through cross-attention. These are valuable methods, but they differ from CDT-II on the dimensions that define our contribution: (i) although iEnhancer-Fusion describes itself as multimodal, the two feature streams it fuses are \textit{both derived from DNA}---sequence semantics and a biophysical property of the same molecule---so its multimodality lies within a single molecular layer, whereas CDT-II integrates two \textit{distinct} molecular layers of the central dogma, genomic context and the observed transcriptome; (ii) their cross-attention fuses representations of the \textit{same} molecule, whereas CDT-II's DNA--RNA cross-attention models transcriptional control, the directed central-dogma relationship itself; and (iii) they output static sequence annotations, whereas CDT-II predicts measured perturbation responses grounded in CRISPRi data. CDT-II's novelty is therefore not the use of cross-attention per se---which, like convolution or self-attention, is a shared computational primitive---but the central-dogma-structured architecture in which each attention mechanism maps onto a defined regulatory relationship and is grounded in perturbation measurements.

\begin{table}[H]
\centering
\footnotesize
\caption{\textbf{Positioning of CDT-II relative to representative method classes.} A head-to-head accuracy benchmark is ill-posed because no other method shares both CDT-II's inputs (genomic context $+$ per-cell transcriptome) and its output (measured held-out-gene perturbation response). Citations for each method appear in the surrounding text.}
\label{tab:positioning}
\renewcommand{\arraystretch}{1.3}
\begin{tabular}{@{}p{2.8cm}p{2.5cm}p{2.4cm}p{1.5cm}p{1.5cm}p{2.6cm}@{}}
\toprule
Method (class) & Input modalities & Output & Cell-state specific & Pert.\ grounded & Interpretability \\
\midrule
Enformer (seq.\ $\rightarrow$ function) & DNA sequence & Cell-type tracks & No & No & Sequence attribution \\
\addlinespace[5pt]
EPBDx\-DNABERT-2; iEnhancer-Fusion (DNA sequence analysis) & DNA only (sequence $+$ breathing) & Sequence labels (TF binding; enhancers) & No & No & Cross-attention over DNA features \\
\addlinespace[5pt]
SCENIC; CellOracle (GRN inference) & scRNA ($+$ motif/ATAC) & Network / \textit{in silico} sim. & Partly & No & Regulon / network \\
\addlinespace[5pt]
GEARS; CPA (perturbation prediction) & scRNA $+$ perturbation label & Post-perturbation expression & Yes & Yes & Limited \\
\addlinespace[5pt]
\textbf{CDT-II (this work)} & \textbf{DNA context ($\pm$57 kb) $+$ per-cell RNA} & \textbf{Held-out-gene log$_2$FC under CRISPRi $+$ interpretable GRN (attention/gradient maps)} & \textbf{Yes} & \textbf{Yes} & \textbf{Directly interpretable per layer: genome regulation (intra-DNA), gene regulation (intra-RNA), and transcriptional regulation (DNA$\to$RNA cross-attention), $+$ gradient attribution} \\
\bottomrule
\end{tabular}
\end{table}

\subsection{Limitations and future directions}

Several limitations should be noted. While in silico knockdown yielded weak correlations ($r \approx 0.07$), gradient-based attribution achieved mean $r = 0.82$ across all five held-out genes, demonstrating that CDT-II captures regulatory relationships when queried appropriately within the training distribution. The 2,361-gene set excludes some known \textit{GFI1B} targets, and all experiments used K562 cells. K562 is currently the only human cell line for which all three layers of the central dogma are jointly available under large-scale CRISPRi perturbation---single-cell RNA together with surface-protein (CITE-seq) measurements, on a deeply characterized genomic and epigenomic background---which makes it the natural anchor for a central-dogma model and underlies the cross-dataset reproducibility filtering central to our approach. CDT-II analyses the DNA and RNA layers (transcriptional control); extending the architecture to the protein layer, so as to model the full central dogma, is the focus of ongoing work. The K562-trained model cannot infer another cell type's baseline \textit{de novo}; applying CDT-II to a new cell type instead requires CRISPRi perturbation data from that cell type, from which the model internalizes its baseline as it does for K562. Because such multi-omic perturbation data are at present available only for K562, generalization to additional cell types awaits comparable atlases and is a primary direction for future work \citep{shalem2014,replogle2022}. Attention weights require careful interpretation \citep{jain2019,wiegreffe2019}; we mitigate this through external validation rather than relying on attention as post-hoc explanation. The relationship of CDT-II to perturbation-prediction and network-inference methods is detailed above (Positioning relative to existing methods).

CDT-II's outputs generate experimentally testable hypotheses: the CTCF binding sites and enhancer elements identified by cross-attention represent specific genomic loci validatable through targeted experiments \citep{fulco2019}, creating a feedback cycle between computational prediction and experimental biology.

The TFRC case study demonstrates this approach's translational potential. The correspondence between gradient predictions and PPMX-T003 clinical and preclinical findings across five of ten functional categories (Table~1) serves as a calibration check that supports the map's biological accuracy. The majority of the 2,361-gene regulatory map constitutes new predictions---for example, the ER stress cluster (PDIA6, SSR2, TMCO1, RTN4, PSMB4) predicts proteostatic disruption as a downstream consequence of TfR1 inhibition, a finding not yet characterized clinically and testable in future studies. This illustrates the AI microscope paradigm: known biology validates the instrument's calibration, after which the full regulatory map can be examined for novel insights.

This analysis can be applied to any perturbation target with Enformer embeddings. All five held-out genes are established or emerging drug targets---\textit{CD52} (alemtuzumab), \textit{CD44} (anti-CD44 antibodies), \textit{TNFSF9} (4-1BB agonists), and \textit{GFI1B} (leukemia therapeutic target)---enabling systematic comparison of pathway consequences across candidates before clinical investment. Extension to multiple cell types via genome-scale perturbation atlases \citep{shalem2014,replogle2022} and integration of protein-level measurements could further sharpen this resolution.

A natural question is whether CDT-II, trained on genetic CRISPRi perturbations, also captures drug-induced transcriptional effects such as those catalogued by the L1000 Connectivity Map \citep{subramanian2017l1000}. The TFRC case study suggests a partial answer: antibody inhibition of TfR1 produces a loss-of-function state resembling CRISPRi knockdown of \textit{TFRC}, which is why CDT-II's knockdown-derived regulatory map aligns with the antibody's clinical and preclinical effects. By the same logic, a small molecule that acts as a specific inhibitor of a gene CDT-II represents should produce downstream consequences approximated by that gene's gradient map. This correspondence, however, is expected to hold only for target-specific loss-of-function agents. Many drugs act through polypharmacology, off-target binding, receptor agonism, or dose- and time-dependent mechanisms, and L1000 signatures further reflect compound identity, concentration, and exposure duration---none of which a CRISPRi-trained model represents. CDT-II therefore cannot, in its current form, predict arbitrary drug signatures; doing so would require training on or benchmarking against compound-perturbation data such as L1000, together with explicit representations of compound identity, dose, and time. We regard this as a concrete and valuable extension: a CRISPRi-grounded regulatory backbone augmented with drug-perturbation data could bridge genetic and pharmacological perturbation, with the loss-of-function targets modelled here serving as mechanistic anchors.

These directions share a common premise: that biological AI should serve not as an oracle delivering predictions, but as an instrument revealing structure. The regulatory networks that govern cellular behavior are distributed across the layers of the central dogma, and no single experiment resolves them all at once. By grounding its architecture in the central dogma, CDT-II brings aspects of that structure into a form a researcher can inspect and test at the bench. It changes the question from ``what does the model predict?'' to ``what does the model reveal?''---the question a microscope has always answered.

\section*{Acknowledgements}

The author thanks the developers of Enformer, the Sanjana laboratory for making the STING-seq dataset publicly available, and the open-source communities behind PyTorch and related tools. Claude (Anthropic) was used for manuscript proofreading and editing.

\section*{Funding}

This work received no external funding.

\section*{Data availability}

The STING-seq v2 dataset is available from GEO under accession GSE171452 \citep{morris2023}. Pre-computed Enformer embeddings, processed training data, and trained model weights are available at \url{https://huggingface.co/datasets/nobusama17/CDT2-data}. ENCODE K562 peak-call datasets are available from the ENCODE portal (\url{https://www.encodeproject.org}). Per-community gene memberships for both attention mechanisms (Supplementary Data~S1) and the corresponding GO and pathway enrichment results are provided in the code repository (see Code availability).

\section*{Code availability}

CDT-II source code is available at \url{https://github.com/nobusama/CDT2}. Supplementary Data~S1 (per-community gene memberships for both attention mechanisms) and the per-community GO and pathway enrichment results are included in the same repository.

\section*{Author contributions}

N.O. conceived the study, developed the model, performed all analyses, and wrote the manuscript.

\section*{Conflict of Interest}

N.O. is the founder of CDT Bio LLC, which is developing applications of the Central Dogma Transformer platform. No other competing interests are declared.

\section*{Figure alt text}

\noindent{\footnotesize
\textbf{Figure 1.} Schematic data-flow diagram in which DNA (Enformer) and per-cell RNA inputs each pass through self-attention blocks, then a DNA-to-RNA cross-attention layer, a Virtual Cell Embedder, and a task head that outputs gene-level log2 fold changes.\par\smallskip
\textbf{Figure 2.} Two line plots: validation learning curves for the 9,335-gene model (plateauing near $r=0.37$) and the 2,361-gene model (rising to $r=0.64$), beside training and validation curves for both models that track each other closely.\par\smallskip
\textbf{Figure 3.} A density scatter of predicted versus observed log2 fold changes hugging the $y=x$ line with a near-zero-centred residual histogram; five per-gene scatter plots of predicted versus observed mean log2 fold change (mean $0.84$); and a profile plot of predicted and observed \textit{GFI1B} trans-effects across genes ranked by observed effect.\par\smallskip
\textbf{Figure 4.} Square DNA self-attention heatmaps over 896 genomic bins with bright vertical bands at the TSS, above five stacked line profiles of mean cross-attention along the same bins, one per held-out gene, each marked with a dashed line at the TSS.\par\smallskip
\textbf{Figure 5.} Per-head bar charts of \textit{GFI1B}'s top RNA self-attention targets showing distinct co-regulatory programs, above a 30-by-30 gene-by-gene heatmap whose dark cells mark the strongest co-regulatory links.\par\smallskip
\textbf{Figure 6.} A horizontal diverging bar chart of \textit{GFI1B}'s top RNA self-attention partners ranked by weight (downstream versus upstream, bold labels bidirectional); a grouped bar chart comparing GO/pathway enrichment between the self-attention and cross-attention communities; and a two-circle Venn diagram (547 self-attention-only, 132 shared core, 33 cross-attention-only) with the 132 core genes listed below by functional category.\par\smallskip
\textbf{Figure 7.} A heatmap of Fisher enrichment for five ENCODE K562 marks across five held-out genes (almost all cells significant), beside a violin plot of cross-attention weight for four bin classes in which the promoter and CTCF distributions sit highest.\par\smallskip
\textbf{Figure 8.} A scatter plot of gradient importance versus experimental effect size for 2,361 genes (top-20 genes in orange, $r=0.86$), beside a five-bar chart of per-gene gradient--experimental correlation with the \textit{GFI1B} bar in red and a dashed mean line at $0.82$.\par\smallskip
\textbf{Figure 9.} A clustered heatmap of absolute Jacobian values (top-100 output genes by top-50 input genes) under \textit{TFRC} perturbation, with a blue-to-red colour scale and dendrogram-grouped blocks marking co-regulated gene modules.
}


\begin{thebibliography}{42}

\bibitem[Aibar et~al.(2017)]{aibar2017}
Aibar,S. \textit{et al.} (2017) SCENIC: single-cell regulatory network inference and clustering. \textit{Nat. Methods}, \textbf{14}, 1083--1086.

\bibitem[Ashburner et~al.(2000)]{ashburner2000}
Ashburner,M. \textit{et al.} (2000) Gene ontology: tool for the unification of biology. \textit{Nat. Genet.}, \textbf{25}, 25--29.

\bibitem[Avsec et~al.(2021a)]{avsec2021bpnet}
Avsec,\v{Z}. \textit{et al.} (2021a) Base-resolution models of transcription-factor binding reveal soft motif syntax. \textit{Nat. Genet.}, \textbf{53}, 354--366.

\bibitem[Avsec et~al.(2021b)]{avsec2021enformer}
Avsec,\v{Z}. \textit{et al.} (2021b) Effective gene expression prediction from sequence by integrating long-range interactions. \textit{Nat. Methods}, \textbf{18}, 1196--1203.

\bibitem[Avsec et~al.(2026)]{avsec2026alphagen}
Avsec,\v{Z}. \textit{et al.} (2026) Advancing regulatory variant effect prediction with AlphaGenome. \textit{Nature}, \textbf{649}, 1206--1218.

\bibitem[Blondel et~al.(2008)]{blondel2008}
Blondel,V.D. \textit{et al.} (2008) Fast unfolding of communities in large networks. \textit{J. Stat. Mech.}, \textbf{2008}, P10008.

\bibitem[Crick(1970)]{crick1970}
Crick,F. (1970) Central dogma of molecular biology. \textit{Nature}, \textbf{227}, 561--563.

\bibitem[Cui et~al.(2024)]{cui2024}
Cui,H. \textit{et al.} (2024) scGPT: toward building a foundation model for single-cell multi-omics using generative AI. \textit{Nat. Methods}, \textbf{21}, 1470--1480.

\bibitem[Dixit et~al.(2016)]{dixit2016}
Dixit,A. \textit{et al.} (2016) Perturb-Seq: dissecting molecular circuits with scalable single-cell RNA profiling of pooled genetic screens. \textit{Cell}, \textbf{167}, 1853--1866.e17.

\bibitem[Dixon et~al.(2012)]{dixon2012}
Dixon,J.R. \textit{et al.} (2012) Topological domains in mammalian genomes identified by analysis of chromatin interactions. \textit{Nature}, \textbf{485}, 376--380.

\bibitem[ENCODE Project Consortium(2012)]{encode2012}
ENCODE Project Consortium. (2012) An integrated encyclopedia of DNA elements in the human genome. \textit{Nature}, \textbf{489}, 57--74.

\bibitem[Eraslan et~al.(2019)]{eraslan2019}
Eraslan,G. \textit{et al.} (2019) Deep learning: new computational modelling techniques for genomics. \textit{Nat. Rev. Genet.}, \textbf{20}, 389--403.

\bibitem[Fauzi et~al.(2025)]{fauzi2025}
Fauzi,Y.R., Nakahata,S., Shimoda,K. \textit{et al.} (2025) Anti-transferrin receptor antibody (JST-TFR09/PPMX-T003) induces ferroptosis in adult T-cell leukemia/lymphoma cells. \textit{Biochem. Biophys. Res. Commun.}, \textbf{756}, 151564.

\bibitem[Fulco et~al.(2019)]{fulco2019}
Fulco,C.P. \textit{et al.} (2019) Activity-by-contact model of enhancer--promoter regulation from thousands of CRISPR perturbations. \textit{Nat. Genet.}, \textbf{51}, 1664--1669.

\bibitem[Gasperini et~al.(2019)]{gasperini2019}
Gasperini,M. \textit{et al.} (2019) A genome-wide framework for mapping gene regulation via cellular genetic screens. \textit{Cell}, \textbf{176}, 377--390.e19.

\bibitem[Gilbert et~al.(2013)]{gilbert2013}
Gilbert,L.A. \textit{et al.} (2013) CRISPR-mediated modular RNA-guided regulation of transcription in eukaryotes. \textit{Cell}, \textbf{154}, 442--451.

\bibitem[Jain and Wallace(2019)]{jain2019}
Jain,S. and Wallace,B.C. (2019) Attention is not explanation. In \textit{Proc. NAACL-HLT}, 3543--3556.

\bibitem[Jassal et~al.(2020)]{jassal2020}
Jassal,B. \textit{et al.} (2020) The reactome pathway knowledgebase. \textit{Nucleic Acids Res.}, \textbf{48}, D498--D503.

\bibitem[Jumper et~al.(2021)]{jumper2021}
Jumper,J. \textit{et al.} (2021) Highly accurate protein structure prediction with AlphaFold. \textit{Nature}, \textbf{596}, 583--589.

\bibitem[Kabir et~al.(2024)]{kabir2024epbd}
Kabir,A., Bhattarai,M., Peterson,S. \textit{et al.} (2024) DNA breathing integration with deep learning foundational model advances genome-wide binding prediction of human transcription factors. \textit{Nucleic Acids Res.}, \textbf{52}, e91.

\bibitem[Kamimoto et~al.(2023)]{kamimoto2023}
Kamimoto,K. \textit{et al.} (2023) Dissecting cell identity via network inference and in silico gene perturbation. \textit{Nature}, \textbf{614}, 742--751.

\bibitem[Kanehisa and Goto(2000)]{kanehisa2000}
Kanehisa,M. and Goto,S. (2000) KEGG: Kyoto encyclopedia of genes and genomes. \textit{Nucleic Acids Res.}, \textbf{28}, 27--30.

\bibitem[Kuleshov et~al.(2016)]{kuleshov2016}
Kuleshov,M.V. \textit{et al.} (2016) Enrichr: a comprehensive gene set enrichment analysis web server 2016 update. \textit{Nucleic Acids Res.}, \textbf{44}, W90--W97.

\bibitem[Linder et~al.(2025)]{linder2025borzoi}
Linder,J., Srivastava,D., Yuan,H., Agarwal,V. and Kelley,D.R. (2025) Predicting RNA-seq coverage from DNA sequence as a unifying model of gene regulation. \textit{Nat. Genet.}, \textbf{57}, 949--961.

\bibitem[Loshchilov and Hutter(2019)]{loshchilov2019}
Loshchilov,I. and Hutter,F. (2019) Decoupled weight decay regularization. In \textit{Proc. ICLR}.

\bibitem[Lotfollahi et~al.(2023)]{lotfollahi2023}
Lotfollahi,M. \textit{et al.} (2023) Predicting cellular responses to complex perturbations in high-throughput screens. \textit{Mol. Syst. Biol.}, \textbf{19}, e11517.

\bibitem[Luo et~al.(2025)]{luo2025ienhancer}
Luo,W., Liu,Y., Yang,L. \textit{et al.} (2025) iEnhancer-Fusion: integrating sequence semantics and DNA breathing dynamics for enhancer identification and strength classification. In \textit{Proc. 2025 IEEE Int. Conf. Bioinformatics and Biomedicine (BIBM)}, Wuhan, China, pp.\ 325--332. doi:10.1109/BIBM66473.2025.11357168.

\bibitem[Morris et~al.(2023)]{morris2023}
Morris,J.A. \textit{et al.} (2023) Discovery of target genes and pathways at GWAS loci by pooled single-cell CRISPR screens. \textit{Science}, \textbf{380}, eadh7699.

\bibitem[Nguyen et~al.(2024)]{nguyen2024}
Nguyen,E. \textit{et al.} (2024) Sequence modeling and design from molecular to genome scale with Evo. \textit{Science}, \textbf{386}, eado9336.

\bibitem[Norman et~al.(2019)]{norman2019}
Norman,T.M. \textit{et al.} (2019) Exploring genetic interaction manifolds constructed from rich single-cell phenotypes. \textit{Science}, \textbf{365}, 786--793.

\bibitem[Novakovsky et~al.(2023)]{novakovsky2023}
Novakovsky,G. \textit{et al.} (2023) Obtaining genetics insights from deep learning via explainable artificial intelligence. \textit{Nat. Rev. Genet.}, \textbf{24}, 125--137.

\bibitem[Ogama et~al.(2023)]{ogama2023}
Ogama,Y. \textit{et al.} (2023) Phase 1 clinical trial of PPMX-T003, a novel human monoclonal antibody specific for transferrin receptor 1, to evaluate its safety, pharmacokinetics, and pharmacodynamics. \textit{Clin. Pharmacol. Drug Dev.}, \textbf{12}, 579--587.

\bibitem[Ota(2026)]{ota2026cdtv1}
Ota,N. (2026) Central Dogma Transformer: towards mechanism-oriented AI for cellular understanding. Preprint at \url{https://arxiv.org/abs/2601.01089}.

\bibitem[Replogle et~al.(2022)]{replogle2022}
Replogle,J.M. \textit{et al.} (2022) Mapping information-rich genotype-phenotype landscapes with genome-scale Perturb-seq. \textit{Cell}, \textbf{185}, 2559--2575.e28.

\bibitem[Roohani et~al.(2024)]{roohani2024}
Roohani,Y., Huang,K. and Leskovec,J. (2024) Predicting transcriptional outcomes of novel multigene perturbations with GEARS. \textit{Nat. Biotechnol.}, \textbf{42}, 927--935.

\bibitem[Saleque et~al.(2007)]{saleque2007}
Saleque,S. \textit{et al.} (2007) Epigenetic regulation of hematopoietic differentiation by Gfi-1 and Gfi-1b is mediated by the cofactors CoREST and LSD1. \textit{Mol. Cell}, \textbf{27}, 562--572.

\bibitem[Shalem et~al.(2014)]{shalem2014}
Shalem,O. \textit{et al.} (2014) Genome-scale CRISPR-Cas9 knockout screening in human cells. \textit{Science}, \textbf{343}, 84--87.

\bibitem[Subramanian et~al.(2017)]{subramanian2017l1000}
Subramanian,A. \textit{et al.} (2017) A next generation connectivity map: L1000 platform and the first 1,000,000 profiles. \textit{Cell}, \textbf{171}, 1437--1452.e17.

\bibitem[Theodoris et~al.(2023)]{theodoris2023}
Theodoris,C.V. \textit{et al.} (2023) Transfer learning enables predictions in network biology. \textit{Nature}, \textbf{618}, 616--624.

\bibitem[Vaswani et~al.(2017)]{vaswani2017}
Vaswani,A. \textit{et al.} (2017) Attention is all you need. \textit{Adv. Neural Inf. Process. Syst.}, \textbf{30}.

\bibitem[Wiegreffe and Pinter(2019)]{wiegreffe2019}
Wiegreffe,S. and Pinter,Y. (2019) Attention is not not explanation. In \textit{Proc. EMNLP-IJCNLP}, 11--20.

\bibitem[Zhao et~al.(2025)]{zhao2025dynamic}
Zhao,W., Solaguren-Beascoa,A., Neilson,G. \textit{et al.} (2025) Multi-modal single-cell foundation models via dynamic token adaptation. \textit{ICLR 2025 Workshop on Machine Learning for Genomics Explorations (MLGenX)}. arXiv:2504.13049.

\end{thebibliography}
\end{document}